%% file: main.tex
\documentclass[10pt,twocolumn,letterpaper]{article}

\usepackage{cvpr}
\usepackage{times}
\usepackage{algorithm}
\usepackage{algpseudocode}
\usepackage{amsmath}
\usepackage{amssymb}
\usepackage{float}
\usepackage{graphicx}
\usepackage{multirow}
\usepackage[numbers,sort]{natbib}
\usepackage[pagebackref=true,breaklinks=true,letterpaper=true,colorlinks,bookmarks=false]{hyperref}
\usepackage{cleveref}

\cvprfinalcopy % *** Uncomment this line for the final submission

\def\cvprPaperID{22} % *** Enter the CVPR Paper ID here

\ifcvprfinal\fi
\begin{document}

\title{Semi-Supervised Learning with Scarce Annotations}

\author{Sylvestre-Alvise Rebuffi\thanks{Authors contributed equally}\hspace{2em} Sebastien Ehrhardt\footnotemark[1] \hspace{2em} Kai Han\footnotemark[1]\\ Andrea Vedaldi \hspace{2em} Andrew Zisserman\\
Visual Geometry Group, University of Oxford\\
{\tt\small \{srebuffi,hyenal,khan,vedaldi,az\}@robots.ox.ac.uk}}

\maketitle

\begin{abstract}
While semi-supervised learning (SSL) algorithms provide an efficient way to make use of both labelled and unlabelled data, they generally struggle when the number of annotated samples is very small.
In this work, we consider the problem of SSL multi-class classification with very few labelled instances.
We introduce two key ideas.
The first is a simple but effective one: we leverage the power of transfer learning among different tasks and self-supervision to initialize a good representation of the data without making use of any label.
The second idea is a new algorithm for SSL that can exploit well such a pre-trained representation.
The algorithm works by alternating two phases, one fitting the labelled points and one fitting the unlabelled ones, with carefully-controlled information flow between them.
The benefits are greatly reducing overfitting of the labelled data and avoiding issue with balancing labelled and unlabelled losses during training.
We show empirically that this method can successfully train competitive models with as few as 10 labelled data points per class.
More in general, we show that the idea of bootstrapping features using self-supervised learning always improves SSL on standard benchmarks.
We show that our algorithm works increasingly well compared to other methods when refining from other tasks or datasets.
\end{abstract}

\input{introduction}
\input{related_work}
\input{method}

\input{experiments}
\input{conclusion}
{\small\bibliographystyle{ieee_fullname}\bibliography{semi}}
\end{document}

%% file: introduction.tex
\section{Introduction}

The success of Deep Learning (DL) in computer vision comes at the cost of collecting large quantities of labelled data.
In many applications, data collection is increasingly inexpensive, but data annotation still involves manual and thus expensive labor.
Semi-supervised learning (SSL) can significantly reduce the cost of learning new models by using large datasets of which only a small proportion comes with manual labels.

When only some data samples are annotated, one can exploit the structure in the data to infer, implicitly or explicitly, the missing annotations.
For example, the class of an image generally does not change if the camera is slightly shifted while maintaining the focus on the same point.
The resulting viewpoint change can be approximated by a deformation or warp of the image.
A change in illumination, which also generally leaves the image class unchanged, can instead be approximated by a linear transformation of the image range.
Hence, an effective way of propagating labels is to enforce consistency in prediction: if one of these perturbations is applied to the image, then the prediction should remain the same. 

%For example, when a camera rotates around an object, it is still easy to point where the object is at every time step without labelling every frame, the object appearance varying only slightly across them.

A simple way to incorporate such invariance while training a deep neural network is to consider image augmentations such as rotations and flips and other perturbations such as Dropout~\cite{srivastava2014dropout}, which is injected directly at the level of the features. Such perturbations
are used  as a form of regularization 
for supervised learning
to avoid overfitting.
The idea is that a perturbed image should maintain the same ground-truth label as the original image.
In the SSL setting, where most labels are unknown, augmentation can be used to encourage the predictions of unlabelled data instances to be `consistent', i.e.~to remain stable across small variations of the input image, regardless of which specific label the data point takes.
If data samples are dense enough, stability to local perturbations may be sufficient to propagate the known labels to the rest of the dataset.
Similar ideas are used in many approaches to semi-supervised classification, including the recent works of~\cite{miyato2018virtual,laine2016temporal,tarvainen2017mean}.
Still, performance degrades quickly with a diminishing number of labelled instances per class~\cite{oliver2018realistic}.

\input{fig-pi-model}

We can explain this performance drop as follows.
Perturbations can bridge some of the gaps between labelled data points, but the effect is ultimately only local.
The ability to transport information across points that are farther apart (since there are fewer of them) depends mainly on the smoothness of the data representation.
For example, the Euclidean distance is essentially meaningless in image space, so that transferring labels between images that are close as vectors of pixels is very ineffective.
On the other hand, if images are first encoded via a representation function in a low dimensional space whose smoothness intrinsically captures some of the relevant invariances, then augmentation can be much more effective.

Unfortunately, simply training a representation such as a deep neural network on a semi-supervised dataset is unlikely to solve the problem. 
In fact, it is difficult to learn effectively from a label-deficient dataset with a small number of data points with known labels and a very large number of data points with unknown labels.
In general, labelled points would work as ``anchors'': since their label is known with certainty, the neural network is forced to fit those points with confidence, so that the information extracted from them can be reliably propagated.
However, when the ratio of labelled vs unlabelled points is very unbalanced, the most likely result is that the neural network would overfit the few labelled points, which would then cease to have an influence on the unlabelled points, thus preventing effective propagation.

We make several contributions to address such shortcomings.
The first contribution is a simple but effective one: we propose to use self-supervision to bootstrap a good representation of the data before running our SSL algorithm.
Self-supervision defines a pretext-task where a full set of labels can be generated.
While these labels are unrelated to the task at hand, we exploit the ability of deep network to \emph{transfer effectively between tasks} and re-use the pre-trained feature to initialize SSL.
Note that the data are the same, it is only the task which is being transferred.

The other contribution is a new SSL algorithm that can leverage the bootstrapped representation well.
The core idea is that, instead of fitting labelled and unlabelled data points simultaneously, we alternate between two phases of optimization.
The information flow between the two phases is carefully controlled to minimize the risk of overfitting.
Namely, during \emph{phase one} we fit the labelled data, but only change the final layer of the representation.
Then, the resulting classifier is used to generate pseudo-labels for \emph{phase two}, where the unlabelled data are fitted.
At this point, the representation is reset to initial state (from the pre-text task) and fine-tuned using the pseudo-labels only.
Finally, the fine-tuned representation is passed back to \emph{phase one} to retrain the classifier and obtain new-pseudo labels.
In this way, the few available data points are only used to fit a small number of model parameters (a classifier layer only), whereas the large unlabelled dataset is used to fine-tune the representation.
We show empirically that this greatly reduces overfitting.

We add other technical contributions to this basic scheme.
The most significant in terms of final performance is inspired by cross-validation, Tri-Training~\cite{zhou2005tri} and weakly supervised localization~\cite{gokberk2014multi}, and amounts to split our dataset in different subsets, considering only part of the unlabelled data at each cycle of the alternate optimization above.

Empirically, we show that our method achieve close to \emph{state-of-the-art} results in a large number of benchmark cases we test.
Furthermore, in addition to showing that our method is effective as a standalone algorithm, we also show two other potential utilities of it. First it is able to refine results obtained from others SSL methods and second it works best among other SSL algorithm when transferring information from one dataset to another.
Finally, further studies on architectures and self-supervised task were carried out to assess their importance in improving SSL accuracy.
Our code can be found at~\url{http://www.robots.ox.ac.uk/~vgg/research/SSL_scarce/}.

%% file: fig-pi-model.tex
\begin{figure*}[h]
\includegraphics[width=0.246\linewidth]{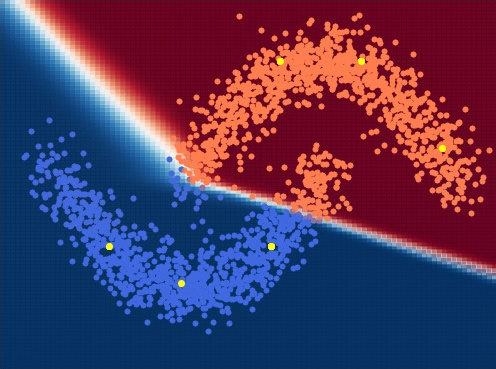}
\includegraphics[width=0.246\linewidth]{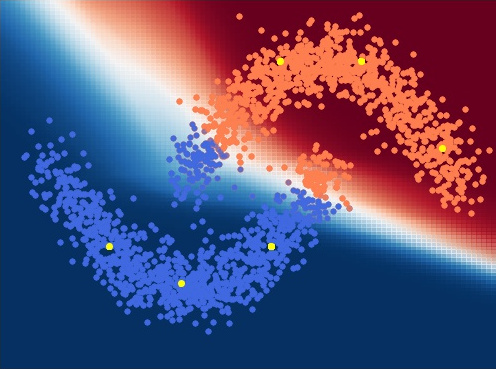}
\includegraphics[width=0.246\linewidth]{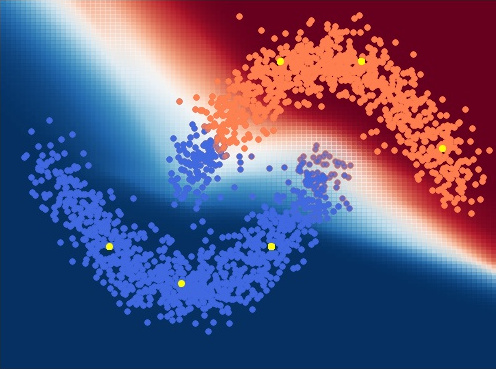}
\includegraphics[width=0.246\linewidth]{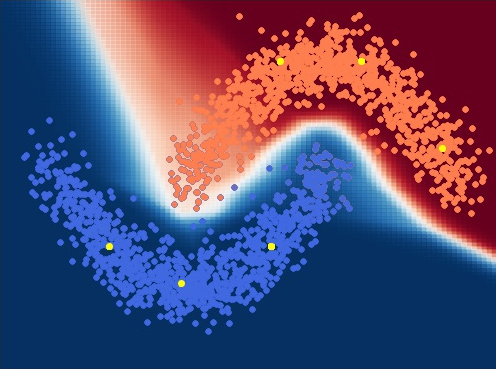}
\caption{\textbf{Consistency loss with the ramp-up strategy.}
The $\Pi$-model~\cite{laine2016temporal} is applied to the `two moons' problem of~\cite{miyato2018virtual} and various steps of the optimization are shown.
At first, the learned network creates a hard border (first panel) based on the labelled anchor points (in yellow).
Then the border is progressively stretched as the consistency loss is enforced more strongly (second and third panels).
Overall, the $\Pi$-model works well to label connected components of the data space. (Darker regions are more confident; best viewed in color.)}
\label{fig:pi-model}
\end{figure*}

%% file: related_work.tex
\section{Related work}\label{s:related}

There exist a vast number of classic works on SSL~\cite{chapelle2006semi} across many disciplines~\cite{fergus2009semi,shi2011semi,teichman2012tracking}.
In the classification context, one common approach is to optimise the conventional cross-entropy loss on the labelled data together with a regularisation term that propagates information to the unlabelled data.
One popular form of `regularisation' is the consistency of predictions to perturbation~\cite{bachman2014learning,sajjadi2016regular}.
\cite{rasmus2015semi} introduced a ladder network that minimises the reconstruction loss between the network outputs from a given sample and its stochastically-perturbed counterpart.
\cite{laine2016temporal} simplified the ladder network into two temporal methods: $\Pi$-Model and Temporal Ensembling.
Both encourage the output of the network for each unlabelled instance to be as similar as possible between different training epochs, by penalising the inconsistency between current network predictions and past predictions.
More recently, the Mean Teacher~\cite{tarvainen2017mean} extended these methods by ensembling over the parameter space:  instead of recalling past predictions they run an exponential moving average over the weights of the network to build a Teacher model.
The Teacher network is then used to enforce consistency in predictions with the main model, called the Student.
Mean Teacher was extended in~\cite{li2019certanity} by a certainty-driven consistency loss that focuses on the unlabelled training samples whose labels are the most uncertain.
Similarly, \cite{miyato2018virtual} considered instead adversarial perturbations that maximise the change in model prediction.
\cite{jackson2019semi} also used this idea together with gradient alignment between labelled and unlabelled data to further improve classification results.

SSL approaches alternative to perturbations include GANs~\cite{springenberg2015unsupervised, salimans2016improved}. \cite{chen2018semi} introduced a memory-assisted deep neural network (MA-DNN) which uses an external memory module to maintain the category prototypes and provide guidance for learning with unlabelled data.
Alternatively, \cite{luo2018smooth} uses a label graph to refine the Mean Teacher's model predictions.
When using graphs, label propagation~\cite{zhu2002learning, li2019generalized, kamnitsas2018semi} is also commonly used. \cite{lee2013pseudo} proposed to iteratively assign pseudo-labels for unlabelled data and pick the high-confidence assignments as training data in the subsequent learning steps, combined with the entropy loss to further regularise  training.
The idea of pseudo labelling is also used alongside with interpolation in \cite{verma2019}.
Tri-training~\cite{zhou2005tri} and its deep network extension Tri-Net\cite{trinet} use co-training~\cite{blum1998combining} to generate three classifiers from three different portions of the data and let them label the unlabelled data. \cite{oliver2018realistic} shows that consistency is generally the most resilient method as the number of labelled instances decreases, but all such methods suffer significantly when this number is decreased past a certain point.

%While it is generally done in segmentation or object recognition tasks~\cite{he2018},
Transfer learning is not often used in SSL. \cite{gidaris2018unsupervised} and \cite{oliver2018realistic} used a pre-trained network to fine-tune it on a restricted part of the labelled set.
Closer to our approach,  \cite{zhou2018when} also combined both SSL algorithms and transfer learning. However, their approach used a pre-trained network from ImageNet\cite{ILSVRC15} to train SSL methods on non-standard datasets for SSL. Our work extends their study to transfer learning from self-supervision and propose a method that surpasses current SSL technique when transferring representation across dataset.

In addition, we notice very recent concurrent works~\cite{Berthelot2019mixmatch,Sohn2020fixmatch} showing promising results. However, none of them focuses on SSL with scarce annotations.

%% file: method.tex
\section{Method}\label{sec:method}

\input{algo-one}

We introduce a SSL algorithm that can work effectively when there are only very few annotated data points, including avoiding supervised pre-training on large labelled datasets.
%Next, we introduce several ideas (\cref{s:self,s:alternate,s:split}) to improve SSL algorithms in this regime and combine them in the new~\cref{ours}.
The algorithm starts by a \emph{preparation phase} (\cref{s:self}, line~1 in~\cref{ours}) that initializes the weights of the model using self-supervision, followed by alternating between two phases (\cref{s:alternate}, lines~5-9 in~\cref{ours}), where in \emph{phase one} a subset of the weights are fitted to the few available labels and in \emph{phase two} the whole model is re-trained from scratch using pseudo-labels on all the data.

\subsection{Transfer learning and self-supervision}\label{s:self}

SSL algorithms risk overfitting the available labelled data points, especially when there is ony a small number of them.
Data augmentation and other forms of regularisation can help to some extent, but their effectiveness is limited.
A more effective solution is to pre-train the model on a much larger dataset, adding to the information contained in the few labelled data points.
Then transfer learning can be used to fine-tune the pre-trained model using the labelled data.

The most common form of pre-training is to use a large, external labelled dataset such as ImageNet.
In this work, however, we focus primarily on the case in which labels are scarce, including in the pre-training phase.
Furthermore, we wish to avoid the use of external data altogether.

%A good representation should project the images into a space that better reflects the semantic similarity between them, suppressing irrelevant differences due to nuisance factors in the imaging process and other intra-class variations, and thus reducing the need for more data augmentation and heavy regularisation.
%In this work, we propose to use transfer learning to initialise the representation.
%The idea is to pre-train the model on an auxiliary dataset or task for which labels are available, and then exploit the ability of deep networks to move from one domain to another effortlessly.

%In our experiments, we consider (mostly) the case in which no external data is available.

In order to do so, instead of using an external labelled dataset for pre-training, we propose to use instead \emph{self-supervision}.
Self-supervision uses a pretext task defined on the available data, both labelled and unlabelled, to bootstrap the model.
The advantage is that self-supervision does not require any label nor external data.

Empirically, we show that self-supervised pre-initialization is very good.
In particular, excellent performance can be obtained by freezing most of the model parameters to their pre-initialized values and use our SSL algorithm to fine-tune only a small subset of them.

\subsection{Alternate optimisation}\label{s:alternate}

Given a pre-initialized model, most SSL algorithms fine-tune it using mini-batches containing a mix of labelled and unlabelled data.
They use a sum of two losses, the first one enforcing the correct classification of the labelled samples and the second one enforcing a form of prediction consistency for the unlabelled samples.
This consistency term usually captures the fact that neighbor data points are likely to have the same label.
Often, a ramp-up is used for the consistency loss~\cite{laine2016temporal,tarvainen2017mean}, so that in the first iterations of training the model focuses on fitting the labelled data.

The effect of this ramp-up can be visualized using the `two moons' of~\cite{miyato2018virtual}, a simple 2D toy clustering problem.
In~\cref{fig:pi-model}, we apply the $\Pi$-model algorithm of~\cite{laine2016temporal} to this dataset.
We see that, before the consistency loss is applied, the network learns a simple boundary that fits the labelled anchor points with tightly.
Then, when the consistency loss is applied, the network stretches the boundary to satisfy the consistency criterion.
This causes the classifier to become less certain near the decision boundary, but more accurate overall.

A drawback of this approach is that, since both losses are optimized jointly, they must be carefully balanced by the choice of appropriate hyper-parameters.
This is particularly true when there are very few label samples, which causes a large imbalance between the two loss terms.
Furthermore, since mini batches are generally evenly split between labelled and unlabelled samples, the few labelled samples will be seen very frequently by the optimizer, which will thus overfit them.

Inspired by incremental learning techniques that try to avoid overfitting on past exemplars while learning new classes~\cite{rebuffi2017icarl}, we propose to disentangle the two losses by alternating the optimisation of labelled and unlabelled data points while maintaining soft constraints between them.
This method reduces the number of hyper-parameters and avoids strong overfitting of the labelled data while still extracting useful knowledge from the labelled set. 

\begin{figure}[h]
\centering
\includegraphics[width=\linewidth]{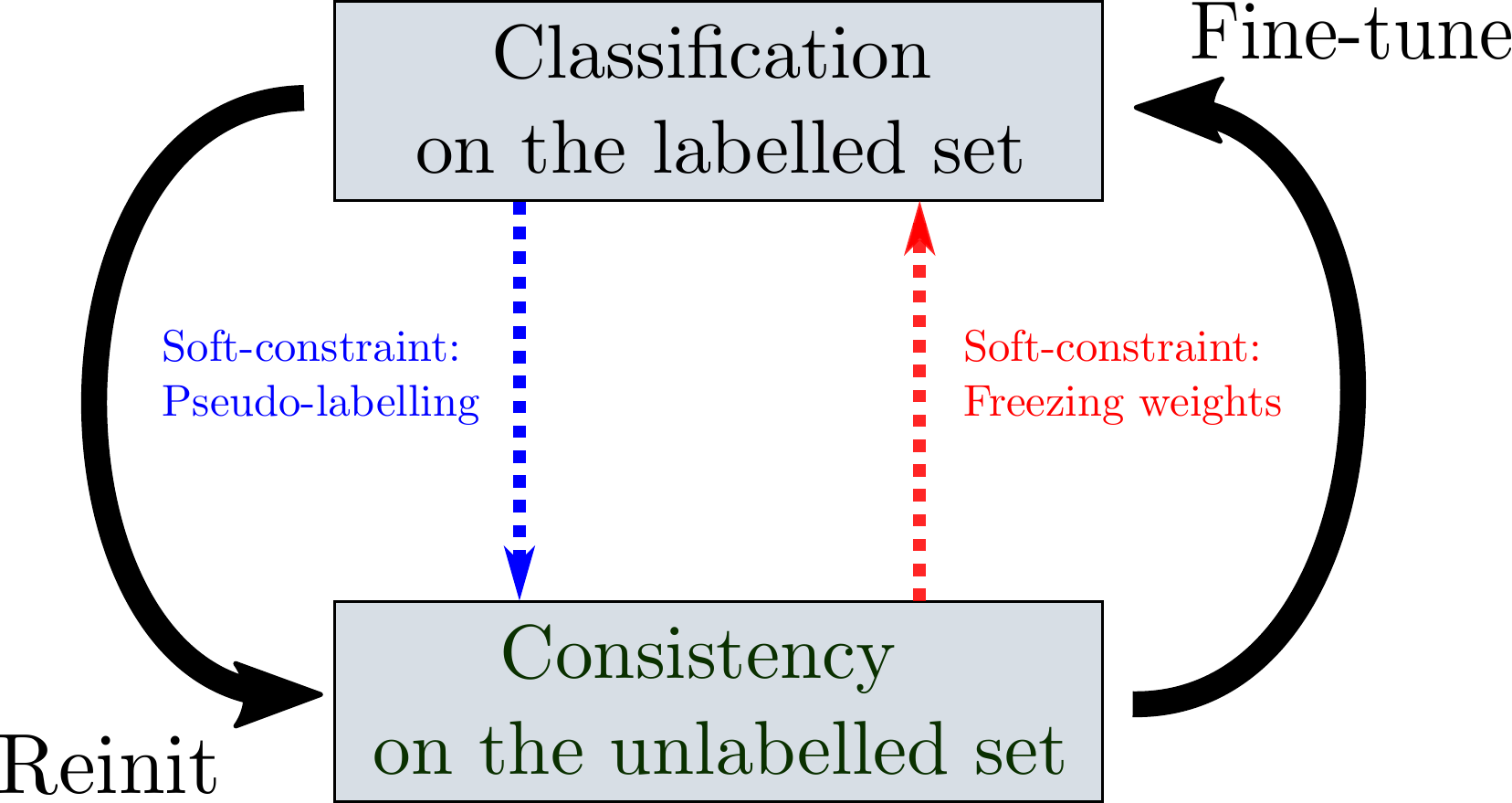}
\caption{\textbf{Overview of our alternating optimisation method.}
Starting from a pre-trained network with a self-supervised learning method we train successively on the labelled and %unlabelled set.
Every time a soft constraint is enforced from previous optimisation.}
\label{fig:method}
\end{figure}

In practice, after the preparatory phase of~\cref{s:alternate} (l.1 in~\cref{ours}), each task is learnt separately in an alternate fashion
(\cref{fig:method})
with a regularisation term that works as a soft constraint borrowing information from the other task.

\paragraph{\emph{Phase one}: fitting the labelled data.}

In the first part of our training, we optimise the cross-entropy loss on the labelled set $\mathcal{L}_{labelled}$ and train the model for a few epochs.
In this part we only fine-tune the final classification layer of the network (l.5 in~\cref{ours}).
This way the labelled samples benefit the most from the feature learnt during optimisation on the unlabelled set while not modifying the intermediate representation.
We then use this trained network to assign soft or hard pseudo-labels to the unlabelled data.
These pseudo-labels are used in the next phase to fit the unlabelled data (l.6 in~\cref{ours}).

\paragraph{\emph{Phase two}: fitting the unlabelled data.}

In the second phase we reset the model with the parameters learnt during the preparatory phase of~\cref{s:alternate} (see l.8 in~\cref{ours}).
Then, we fine-tune the whole architecture on the unlabelled set using a loss $\mathcal{L}_{unlabelled}$ that is a weighted average of a term $\mathcal{L}_{pseudo}$ fitting the pseudo-labels estimated in the first phase, and a second temporal consistency term $\mathcal{L}_{temp}$, borrowed from the $\Pi$-model of~\cite{laine2016temporal}. The idea of the second term is to match the probability distribution $p_i^{t-1}$ obtained from the network for a sample $x_i$ at epoch $t-1$ with the probability $p_i^t$ assigned at iteration $t$ to the same sample. We use the KL-divergence between $p_i^{t-1}$ and $p_i^t$ during \emph{phase two}. (l.9 in~\cref{ours}). The loss $\mathcal{L}_{pseudo}$, besides being a soft constraint tying \emph{phase two} to \emph{phase one}, can also be viewed as another consistency condition~\cite{zhou2018when} that also helps reduce the entropy on the final prediction.
%

%
%The final loss $\mathcal{L}_{unlabelled}$ is a weighted sum of the temporal consistency loss $\mathcal{L}_{temp}$ and the cross entropy loss using the pseudo- labels $\mathcal{L}_{pseudo}$. We use equal weights of 0.5 for each loss in the overall loss.

\paragraph{Information flow.}

Note that the model is reinitialised every time \emph{phase two} is entered.
This is done to avoid getting stuck in local minima.
Hence, the information that flows
(\cref{fig:method})
from \emph{phase one} to \emph{phase two} is only the pseudo-labels on the unlabelled data, whereas the information that flows from \emph{phase two} back to \emph{phase one} are the weights of the network fine-tuned on the unlabelled data.

While these choices may seem ad-hoc, they constitute the primary reason why our algorithm is able to work well in a low-labelled-data regime by avoiding overfitting.
Thanks to this technique, the algorithm focuses on fitting one loss at a time while still allowing information to be exchanged between the two subtasks.
We show in~\cref{sec:exp} this approach robustly improve the quality of the overall classifier at every step.

\subsection{Dataset split}\label{s:split}
One potential pitfall of alternating optimisation is the drift that may be induced by  training for a long time on the unlabelled data.
Inspired by cross-validation and co-training~\cite{blum1998combining}, we also suggest to split the dataset into several parts.
However, differently from~\cite{blum1998combining} we do not train different models from each split.
Instead, at the beginning of each training cycle we randomly choose two-thirds of the unlabelled training data to learn from (l.7 in~\cref{ours}).
The rest of the unlabelled data is held out to reduce the risk of overfitting to it in the current training cycle.
%\todo{Is this true given that the network is restarted anyway?}
This way, the held-out samples are more likely to have their labels swapped during pseudo-label generation in the next cycle.
Roughly two-thirds of this ``fresher'' held-out data will be used to fit the model in the next cycle.
The benefits of this reshuffling procedure diminish as training nears completion, so all the data is used in the final few phases of training.
In~\cref{sec:exp} we demonstrate empirically the benefits of this approach.

%\subsection{Losses on the unlabelled set}\label{s:unsup-loss}
%
%As state above, phase 2 fits the unlabelled set using the cross-entropy loss on the pseudo-labels assigned to every unlabelled sample $x_i$ after phase 1 completes.
%This loss, besides being a soft constraint tying phase 2 to phase 1, can also be viewed as another consistency condition~\cite{zhou2018when} that also helps reduce the entropy on the final prediction.
%
%We also borrow the idea of using a temporal consistency loss from the $\Pi$-model of~\cite{laine2016temporal} due to its excellent performance.
%The idea is to match the probability distribution $p_i^{t-1}$ obtained from the network for a sample $x_i$ at epoch $t-1$ with the probability $p_i^t$ assigned at iteration $t$ to the same sample.
%We use the KL-divergence between $p_i^{t-1}$ and $p_i^t$ during phase 2.
%
%The final loss $\mathcal{L}_{unlabelled}$ is a weighted sum of the temporal consistency loss $\mathcal{L}_{temp}$ and the cross entropy loss using the pseudo- labels $\mathcal{L}_{pseudo}$. We use equal weights of 0.5 for each loss in the overall loss.

% AV: sounds out of place here:
%We note here that it is important to use data augmentation during the reassignment process as there might be overfitting on the central crop of the image

%% file: algo-one.tex
\begin{algorithm}[h]
\caption{Proposed Alternative Optimisation Algorithm}\label{a:one}\label{ours}
\begin{algorithmic}[1]
\State \textbf{Preparation phase:}
    \State Train a self-supervised method on the whole dataset and freeze the first blocks' weights. Replace the last layer with one dimensioned for the classification task. The trainable weights now form a network $N_t$
\State \textbf{Main Loop:}
\For{$i \in \{1,\dots,N\}$}
    \State \textbf{Supervised-training}: Fine-tune $N_t$ classification layer on the labelled subset with cross-entropy loss $\mathcal{L}_{labelled}$.
    \State \textbf{Labels assignement}: Use $N_t$ to assign a label $y_i$ to each unlabelled sample $x_i$.
  \State \textbf{Dataset split}: Create a training set $T_i$ from a random split of the unlabelled data.
  \State \textbf{Restart}: Reassign $N_t$ to the weights extracted from the preparation phase.
  \State \textbf{Unsupervised-training}: Train $N_t$ on the unlabelled metaset $T_i$ with consistency and pseudo labelling loss: $\mathcal{L}_{unlabelled} = 0.5\mathcal{L}_{temp}+ 0.5\mathcal{L}_{pseudo}$.
  %\State Define the loss $L=0.5 \times L_{cross\_entropy} + 0.5 \times L_{self\_distillation}$
      %\State Finetune the base network $N_0$ on $T_i$ using the loss $L$
      %\State Finetune the classifier layer on the metaset and assign new labels
\EndFor
%\State
%\State \textbf{Finetune Loop:} Do a few label assignment cycles on the whole dataset without any self distillation and metaset feedback
\end{algorithmic}
\end{algorithm}

%% file: experiments.tex
\section{Experiments}\label{sec:exp}
%% Would be better with pictures here
\begin{figure*}[h]
\centering
\includegraphics[width=0.33\linewidth]{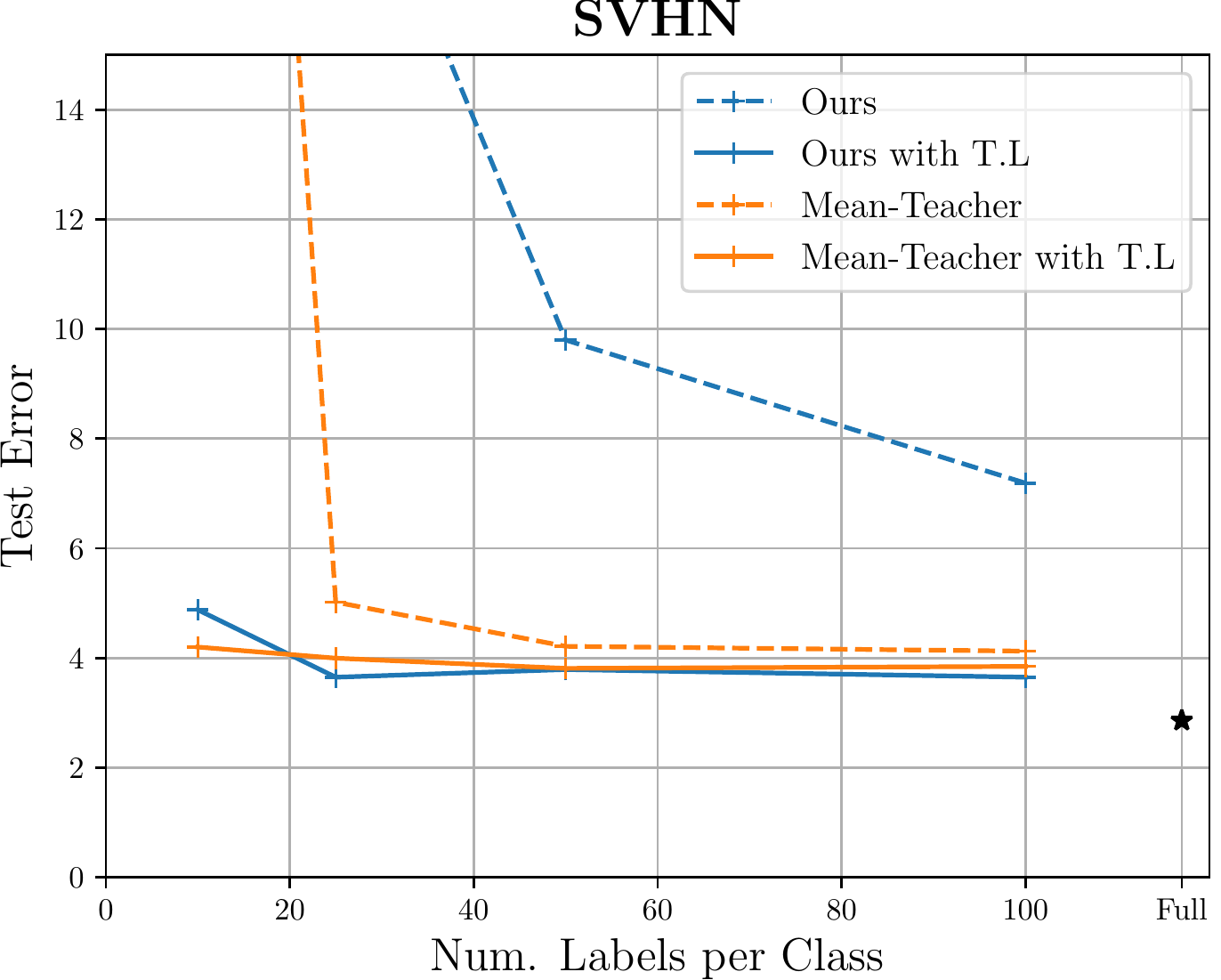}
\includegraphics[width=0.33\linewidth]{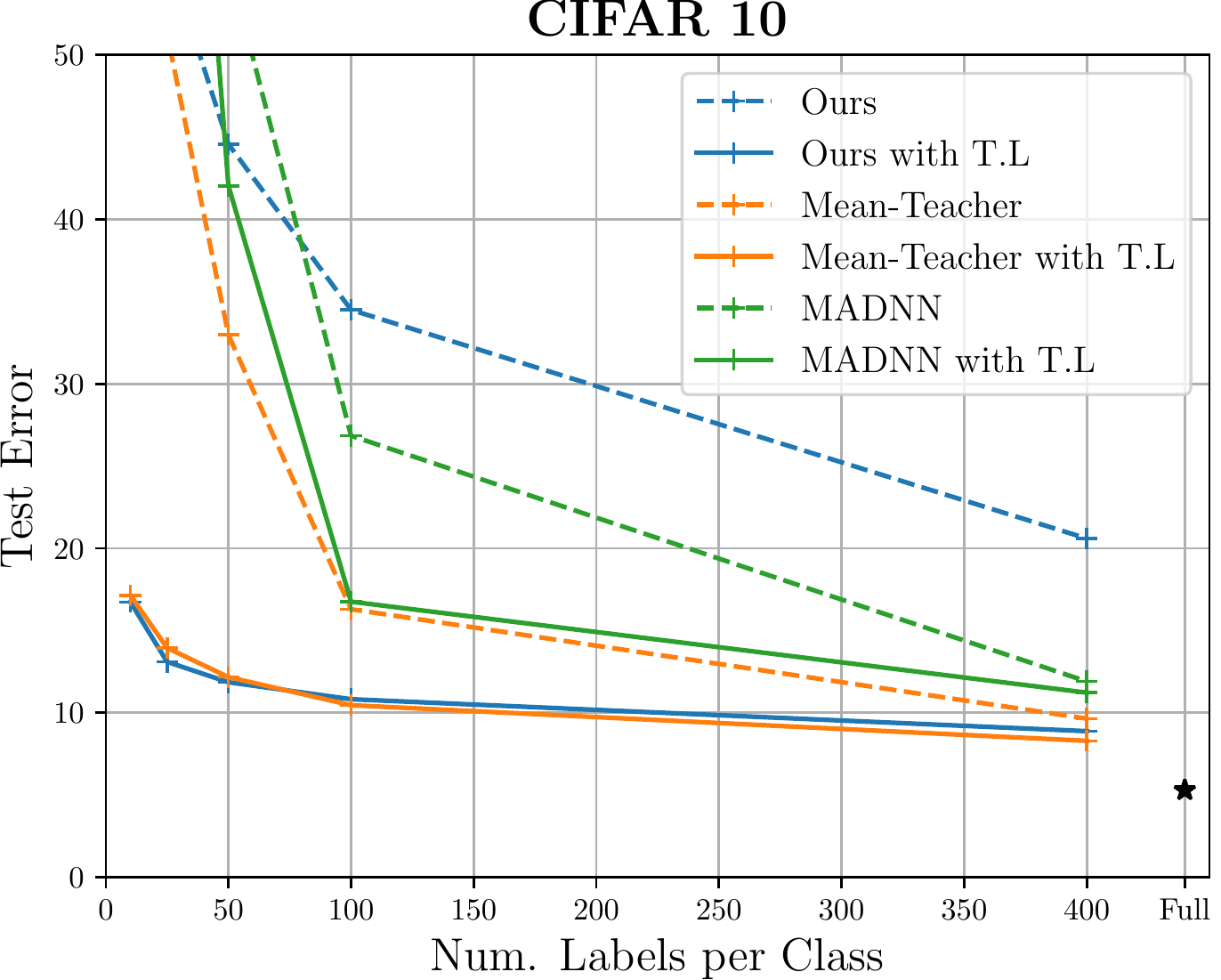}
\includegraphics[width=0.33\linewidth]{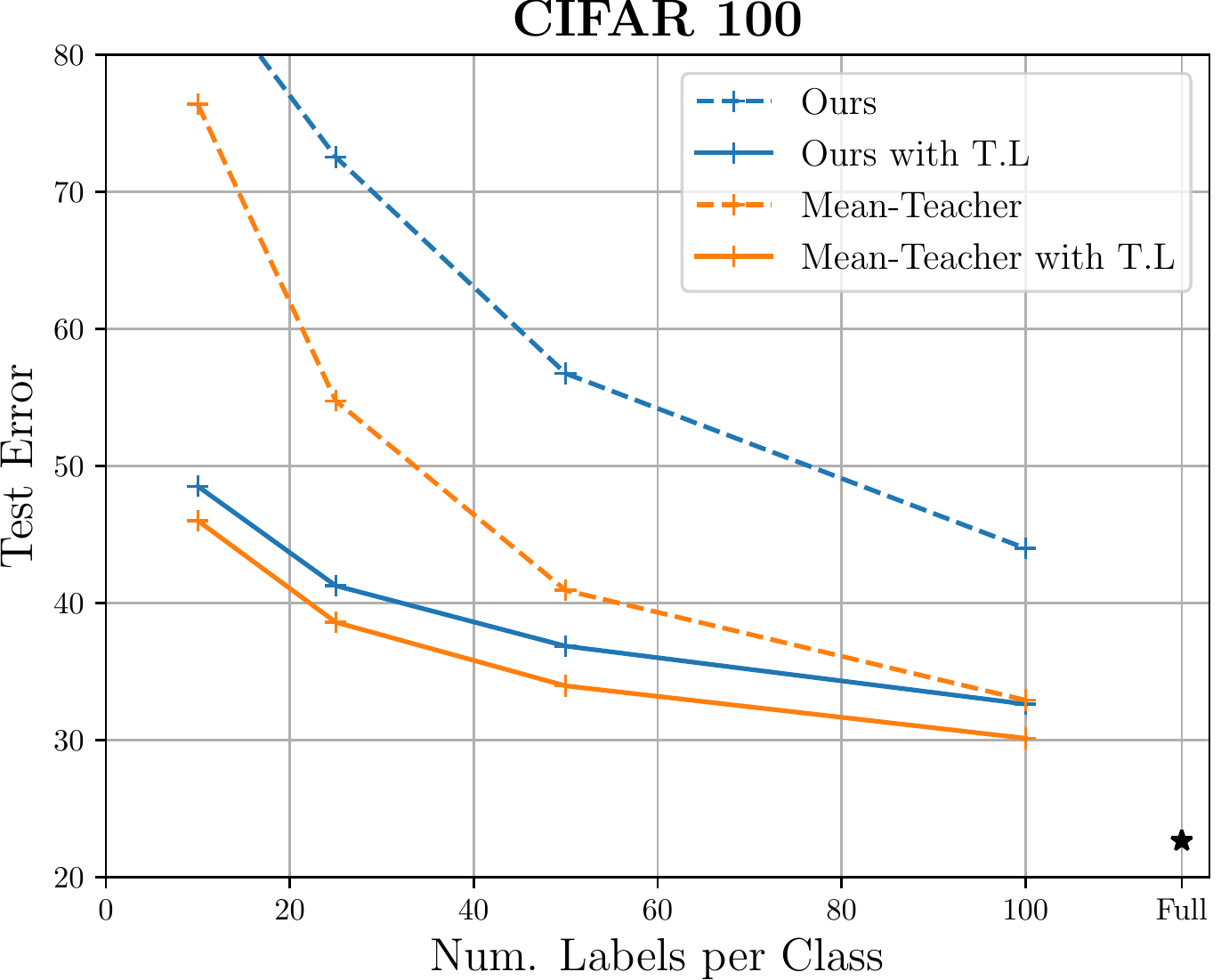}
\caption{\textbf{Impact of Tranfer learning for SSL classification benchmark.} For each dataset we vary the number of labelled data per class in ${10,25,50,100}$ with an additional experiment on CIFAR-10 with 400 labels per class. We note that for every method, models benefit from self-supervision. The `with T.L' indicates a model pretrained with RotNet\cite{gidaris2018unsupervised}. In addition we can see that our method is getting competitive results with Mean Teacher. The star point `*' denotes our network trained with supervision only on the full training set. The shown results are the average of 10 runs per setting.}
\label{fig:comp}
\end{figure*}
\paragraph{Datasets.} Following \cite{chen2018semi} we mostly use standard image classification tasks suitably modified to the semi-supervised context. Each dataset is randomly split in two sets: one small set containing images with their corresponding labels evenly divided among classes and another set with the rest of the image without annotations. For each dataset we evaluated our method on the publicly available test set. We used the same datasets as \cite{chen2018semi}:

\noindent \textbf{SVHN}~\cite{netzer2011svhn}. A Street View House Numbers dataset including 10 classes
(0-9) of coloured digit images from Google Street View. The classification task
is to recognise the central digit of each image. We use the format-2 version
that provides cropped images sized at $ 32\times32$, and the standard 73,257/26,032
training/test data split. We centered and scaled all images RGB channels based on the extracted dataset statistics and augmented the data with random cropping.

\noindent \textbf{CIFAR-10}~\cite{krizhevsky09cifar}. A natural images dataset containing 50,000 training and 10,000 test image samples from 10 object classes. Images have a $32\times32$ resolution and are evenly divided among classes. We again pre-processed the data with centering and scaling and augmented the data with random cropping and flipping.

\noindent \textbf{CIFAR-100}~\cite{krizhevsky09cifar}. A dataset (with the same image size as CIFAR-10) containing 50,000/10,000 training/test images from 100 more fine-grained classes. We used the same pre-processing and data augmentation as CIFAR-10.

\paragraph{Baselines.} We used Mean-Teacher~\cite{tarvainen2017mean} as the main baseline to experiment alongside with our proposed method. Additionally we also used MA-DNN~\cite{chen2018semi} on a more restricted set of experiments. For both we used publicly available code provided by the authors\footnote{\url{https://github.com/CuriousAI/mean-teacher}; \url{https://github.com/yanbeic/semi-memory}}.

\paragraph{Implementation details.} Unless stated otherwise,  we used the ResNet-18~\cite{he2016deep} model with then average pooling and skip connections. We chose RotNet~\cite{gidaris2018unsupervised} as the self-supervision method as it is easy to implement and integrate in any pipeline. For all networks we trained this proxy task on 200 epochs with a step-wise decaying schedule. A more thorough study is done in \cref{ssec:self}. All networks are trained using Nesterov accelerated gradient method~\cite{nesterov1983method}. We use the same hyper-parameters when training with and without transfer learning, which have every time been tuned on the specific task.

\subsection{Impact of self-supervision}
We first compare Mean-Teacher and our algorithm on standard classification benchmarks. For every dataset we trained the network with 10/25/50/100 labels per class, with an extra experiment of 400 labels per class on CIFAR-10 as is common practice~\cite{laine2016temporal,tarvainen2017mean,chen2018semi}. In every experiment we keep the same split of the data for each method. In the default case, models are trained from scratch. For every method we also used transfer learning from self-supervision, with only the last two blocks and the last classification layer used for semi-supervised training (Legends containing `with T.L.'). Results are shown in \cref{fig:comp}. 

This experiment clearly demonstrates the importance of transfer learning as the number of labelled data per class decreases. For instance, on CIFAR-10, transfer learning from self-supervision used with semi-supervision makes an improvement of over 50 points in classification error for both our method and Mean Teacher. On SVHN and CIFAR-10 the performance from 100 labels per class to 10 labels per class decreases by less than 5 points in error. Performances on CIFAR-100 dropped by a larger amount when having fewer labels due to the higher complexity of the dataset, although the performance deteriorate noticeably more slowly when using self-supervision pre-training. Overall this means that self-supervision enables networks to train with competitive results in the extremely low data setting  where standard SSL methods fail. 
% TODO: change this
We also emphasise that we only use the dataset provided and did not use any regularisation method and very limited data augmentation. This is particularly crucial and demonstrates that a carefully chosen self-supervision method can be sufficient to extract very good features which are sufficient to train good models with few labels.

For every dataset we see in \cref{fig:comp} that our method and Mean-Teacher obtain similar performance when they both use transfer from self-supervised networks. However our method needs a more careful initialisation as its performance dropped for models trained from scratch. This can easily be understood, as fine-tuning on the labelled set would be more efficient with either an increased number of labelled data or better intermediate representation. In the rest of the paper we call a `good' representation of an image, a representation from which it is easier to discriminate different classes. %Supposedly this representation is better capturing the distinctive features of every classes.

In addition, we show that self-supervision can also improve the state-of-the-art performance of a more recent SSL method. To this end we use MA-DNN~\cite{chen2018semi} with the 10-layers model proposed in their paper. Again we used RotNet~\cite{gidaris2018unsupervised} as the self-supervision method and only fine-tune the last block and linear layer on the SSL task. We see in \cref{fig:comp} that again, transfer from self-supervision consistently achieves the best performance, improving their \emph{state-of-the-art} results on CIFAR-10 from 11.9\% error to 11.1\% using 400 labels per class. With the hyper-parameters given in their paper, we demonstrate again that under equivalent settings, self-supervision is beneficial to SSL. 
% TODO: maybe dont include
Finally we found that MA-DNN was more sensitive to the number of labels per class. This means that building a reliable template of every class requires more annotations, while we observed that consistency based methods are seemingly showing better robustness overall. 

\subsection{Ablation study}
To assess the impact of every component of our method we also conducted an ablation study on the CIFAR-10 SSL task with 10 labels per class. 

\begin{figure}[h]
\centering
\includegraphics[width=0.9\linewidth]{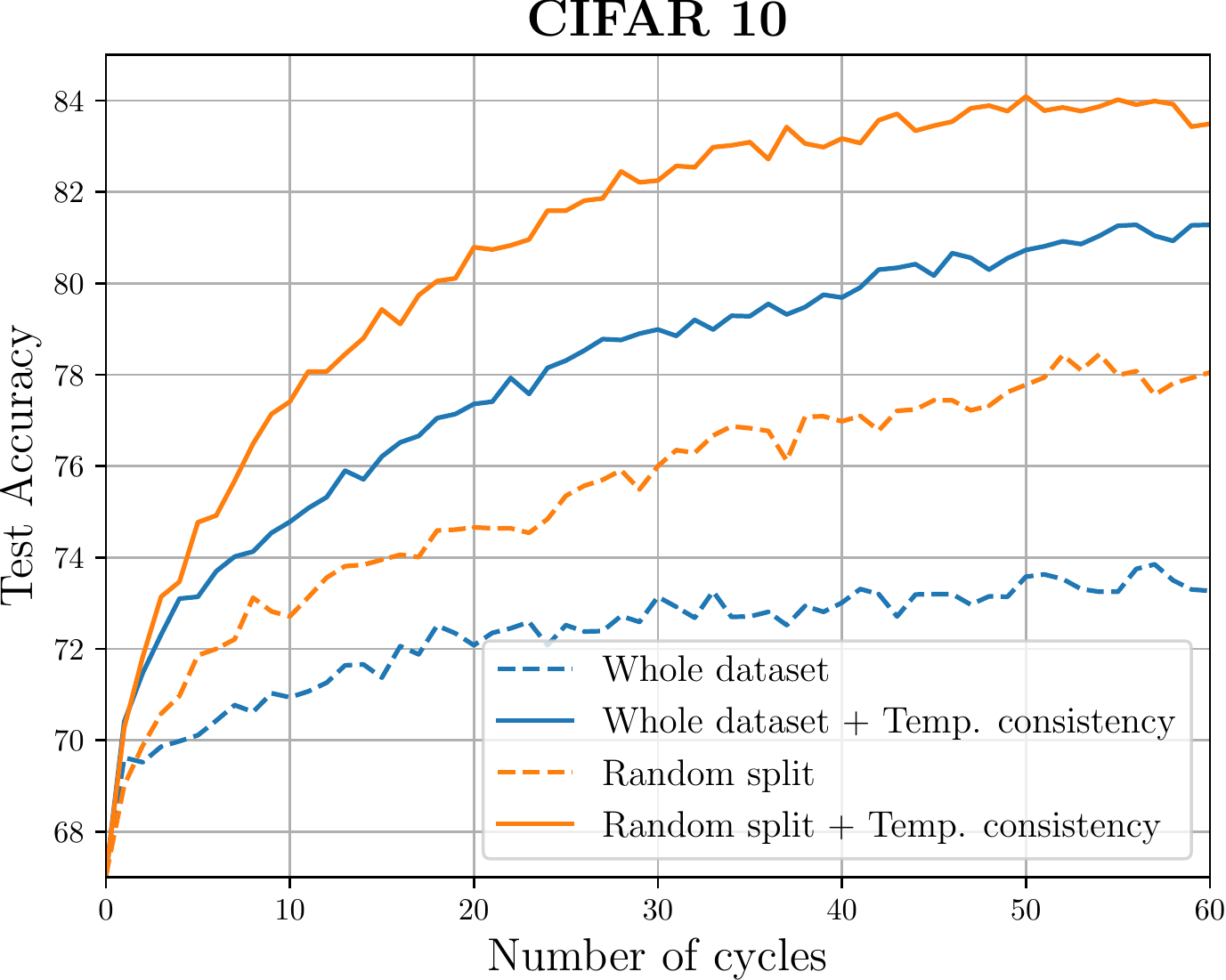}
\caption{\textbf{Ablation study.} Starting from a pre-trained model, our alternating optimisation method makes steady progress over cycles. Dataset random splitting is also effectively speeding-up training. Default case uses $\mathcal{L}_{unlabelled} = \mathcal{L}_{pseudo}$. `Temp. consistency' indicates that we use both $\mathcal{L}_{temp}$ and $\mathcal{L}_{pseudo}$.}
\label{fig:ablation}
\end{figure}

We first measure the gain of using random splitting of the unlabelled set through training cycles as opposed to using the whole set. We observe in~\cref{fig:ablation} that random dataset splitting allows much faster progress in the course of training. When only using pseudo-labelling on the unlabelled set and no temporal consistency, this technique also leads to a better optimum.

Then we analyse the effect of the consistency in the training cycle. The dashed curves correspond to cases where the temporal consistency is not used. It means that, when training on the unlabeled set, we only optimise the cross-entropy on the pseudo-labels previously assigned during the training phase on the labelled set. Interestingly, we also note that the performance is still increasing over cycles even without an explicit consistency loss. In fact, this corroborates the statement of \cite{zhou2018when}, that the  assigned pseudo-labels act as an implicit consistency across samples (instead of the explicit consistency across augmented versions of the same sample) which improves the representation for the next training phase on the labelled set.

Finally we can observe for the different settings that the performance is steadily increasing over cycles. The training indeed does not stall after a few cycles but instead keeps improving the feature representation over the course of alternating optimisation cycles. Alternating optimisation therefore refines the results every cycle with each optimisation helping the next one to reach a better optimum.
In addition, we conduct another ablation experiment skipping re-initialization (line 8 in~\cref{ours}). Compared to the two-phase algorithm with re-initialization, the accuracy drops by 12 points on CIFAR-10 and even 39 points on CIFAR-100. When re-initialization is turned on, 10$\times$ more samples switch labels every assignment cycle, thus avoiding local minima. %Indeed, the re-initialization forces the data representation to stay closer to the initial representation, which is either coming from self-supervised learning or pretraining on another dataset. 
To the best of our knowledge, other existing SSL methods do not have any specific mechanism against representation forgetting when doing transfer learning. 

\subsection{A refining SSL algorithm}\label{ssec:refine}
In~\cref{fig:comp} we note that our alternating training method is particularly efficient when using a good representation rather than starting from scratch. Hence we hypothesise that our method could make an even bigger difference when starting from a better network representation. One way of doing it is to build our method on top of others in order to refine their results. To validate this hypothesis, we apply our method as a refinement on CIFAR-10 and CIFAR-100 datasets for networks pre-trained with Mean-Teacher~\cite{tarvainen2017mean}. It should be noted that, as the starting features are already competitive, we do not use the dataset random splitting in our training procedure and we choose soft labels assignment in l.6 of~\cref{ours}. 
%% Refinement Experiment for 10 labels per class mean teacher and not
\begin{table}[t]
\centering
\begin{tabular}{lccc}
	& CIFAR-10 & \multicolumn{2}{c}{CIFAR-100}
\\ 
  Labels per class & 10 & 10 & 25 \\
\hline
\hline
    MT & 32.4 & 23.6 & 45.7\\
\hline
	MT + Self-Sup  & 86.4 & 53.7  & 60.7
\\
\hline
	MT + Self-Sup + Refinement & \bfseries 87.7 & \bfseries 58.3 & \bfseries 63.9

\end{tabular}

\caption{\textbf{Our method applied as a refinement of Mean Teacher (`MT').} For different numbers of labels per class our method always improve the testing accuracy by a big margin.}
\label{tab:refinement}
\end{table}

In~\cref{tab:refinement}, we see that in every case our method used as refinement consistently improves the test accuracy. The synergy of alternating training is enabling the network to gain as much as 4.6 points in test accuracy on CIFAR-100 and 1.3 points on CIFAR-10. The refinement technique combined with self supervision allows us on CIFAR-10 to almost match previous \emph{state-of-the-art} results with methods trained with 400 labels per class ($11.9$\% reported in \cite{chen2018semi}) whereas we only used 10 labels. These results confirm our previous hypothesis and motivate us to see the impact of increasingly better representations in the next section.

\subsection{Transfer learning from different tasks}
So far we have only used the dataset at hand to extract our intermediate representation with self-supervised learning. In this section we use a different type of transfer learning used in \cite{zhou2018when}: we learn our representation on a different classification dataset. To this end, we use networks pretrained on ImageNet~\cite{ILSVRC15}. As is done in the other sections, we then fix the first two blocks of our ResNet-18 network and train the rest of the weights on the hardest semi-supervised learning task of 10 labels per class using either Mean Teacher or our alternating optimisation method (see \cref{sec:method}). We experimented with CIFAR-10 and CIFAR-100 with pretraining on a downsampled $32\times32$ version of ImageNet, and also MIT Indoor-67~\cite{quattoni2009recognizing}, and `Places10' where we only kept 10 classes from Places~\cite{zhou2014learning} using a pretraining on the standard version of ImageNet. Results are shown in~\cref{tab:transfer} where we perform 10 runs per setting. Fine-tuning on the full set is given as a reference. 
%To this end, we train our network on a downsampled $32\times32$ version of ImageNet~\cite{ILSVRC15}
%% Transfer Learning
\begin{table}[t]
\centering
\setlength{\tabcolsep}{0.3em}
%\small
\begin{tabular}{lcccc}
	& C-10 & C-100 & P-10 & MIT-67
\\
\hline

	Fully Sup.  & 96.6 & 82.5 & 83.3 & 75.5
\\
\hline
\hline
	Labelled Set  & 82.5$\pm$1.0 & 62.8$\pm$0.6 & 66.4$\pm$1.7  & 55.1$\pm$0.8
\\
\hline

	M.T.\cite{tarvainen2017mean}  & 88.0$\pm$1.5 & 63.3$\pm$0.7 & 72.8$\pm$1.2 & 58.8$\pm$1.3
\\
\hline
	Ours & \bfseries 93.5$\pm$0.4 & \bfseries 69.2$\pm$0.4 & \bfseries 75.8$\pm$1.7 & \bfseries 63.0$\pm$0.7

\end{tabular}
\caption{\textbf{Transfer Learning: Test accuracy on the target task.} Transfer is from training on ImageNet. We compare our method with Mean Teacher  (M.T.) and supervision on the labelled set only (`Labelled set') for 10 labelled sample per class. Our method consistently outperforms both method on this task and achieves results nearly as good as the one obtained by training on the full labelled set.}
\label{tab:transfer}
\end{table}

We note that our method consistently outperforms Mean Teacher for such scenarios. This result corroborates our assumption in the previous section that our method performs increasingly well with better representation. In fact, as also noted in \cite{oliver2018realistic}, since ImageNet has redundant classes with the datasets, we can consider the representation learnt from ImageNet as an upper bound. Thus it is not entirely surprising to see our method obtaining nearly as good results as if it were using the complete set with 93.12\% accuracy compared to 96.56\% when fine-tuning on the full labelled set on CIFAR-10. As our method aims to disentangle the two losses by using distillation to preserve the effect of the very reduced labelled set, its effect is magnified when using a representation as good as the one learnt on ImageNet. In contrast, techniques like Mean Teacher overfit on a small labelled set, which in turns will lead to the degradation of the representation coming from ImageNet.

\subsection{Evaluating representation quality with SSL}\label{ssec:arch-search}
In the context of SSL for images classification, practitioners usually use shallow networks. If some works \cite{verma2019,tarvainen2017mean} also experimented with deeper and wider architectures, then those were generally applied when the number of labels at hand was large enough (4,000 on CIFAR-10, $\sim$100,000 on ImageNet). So far in this work, we have shown that it is possible to train a deeper architecture by using transfer-learning from self-supervision. In practice, what matters in our context is the quality of the extracted intermediate representation that we will train our model from. 

In this section we experiment with two other architectures and use SSL to measure the quality of their representation. As is done in~\cite{kol2019}, we compare with one fully-convolutional model and a  RevNet\cite{gomez2017reversible} model. We chose the fully convolutional 10-layers/3-blocks model from~\cite{laine2016temporal} which we named `TempEns'. We provided our own RevNet implementation named `RevNet-18' aimed to match the ResNet-18 we used in this work. More specifically, the RevNet-18 has four main blocks and uses the same type of downsampling mechanism as our ResNet-18. We pre-trained all models with RotNet and trained our SSL method on CIFAR-10 with different numbers of annotated instances. For all architectures the first two blocks of the networks are fixed during SSL training, as we found it was giving best results for every architecture. We report results in~\cref{tab:arch}.
%% Transfer Learning
\begin{table}[t]
\centering
\begin{tabular}{lccc}
$N$& TempEns\cite{laine2016temporal}&RevNet-18\cite{gomez2017reversible} & ResNet-18\cite{he2016deep}\\
\hline
10	& 18.4$\pm$1.9  & 20.0$\pm$2.4 &\bfseries 16.4$\pm$2.6
\\
\hline
50	& 16.3$\pm$0.9  & 13.7$\pm$0.7 & \bfseries 11.9$\pm$0.4
\\
\hline
100	& 14.8$\pm$0.6  & 12.2$\pm$0.5 & \bfseries 10.8$\pm$0.4
\\
\hline
400	& 12.8$\pm$0.2  & 10.5$\pm$0.4 & \bfseries 8.9$\pm$0.2
\\
\hline
Full & 6.7 & 5.8 & 5.3
%	TempEns\cite{laine2016temporal} & 16.8 $\pm$   & 15.6 & 13.7 & 12.2 &6.%7
%\\
%%\hline
%   % WideResNet\cite{zagoruyko2016wide} &&&&
%%\\
%\hline
%    RevNet-18\cite{gomez2017reversible} & 42.7 & 13.7 & 11.9 & 10.2 & 5.8
%\\
%\hline
%	ResNet-18\cite{he2016deep} & \bfseries 16.4 $\pm$ 2.6 & \bfseries 11.9 $\pm$ 0.4 & \bfseries 10.8 $\pm$ 0.4&  \bfseries 8.9 $\pm$ 0.2 & \bfseries 5.3

\end{tabular}
\caption{\textbf{Model variation.} Test error on CIFAR-10 with different number of labels per class `$N$'. In all cases networks, were pre-trained with RotNet~\cite{gidaris2018unsupervised}. First two block are then frozen while the rest of the network is fine-tuned on the classification task. `Full' indicates that we used the full training set with a fully supervised method. All methods uses the same hyper-parameters.}
\label{tab:arch}
\end{table}

We found that the ResNet-18 has better performance over all methods. TempEns surprisingly remains very competitive for all cases with less weights trained than both methods, with good robustness to extreme cases. While RevNet is almost always able to reach ResNet-18 level of performances. With such sparse data, since RevNet has the most parameters to fine-tune it is thereby the most prone to overfit. Overall, the RevNet results slightly contrasts with the one found by \cite{kol2019}, where a RevNet representation surpasses ResNet when training with rotation. However, we note that their methods of evaluation differs significantly from ours. These results emphasise that while K-means neighbors or full supervision can be used to evaluate self-supervised learning~\cite{zhang2019,kol2019}, SSL could also be  used as a way to evaluate the intermediate representation learnt from self-supervised learning algorithms.

\subsection{Self-supervision performance vs classification accuracy}\label{ssec:self}
In general, different self-supervision tasks would lead to different intermediate representation. However one could also wonder how much solving the proxy task could help to find a representation that would be best suited for the classification task. Intuitively, heavy fine-tuning on the proxy task could lead to a more specialized network which would be harder to adapt to the classification task. To verify this, we studied the rotation accuracy impact on the final SSL performance. Our self-supervised method was trained for 200 epochs with a step-wise decaying learning rate starting from 0.1 at epochs 60, 120, and 160. We saved snapshot of the network before decaying the learning rate and measure the performance reached by our method on the CIFAR-100 classification task with 10 labelled instances per class. In addition, we also trained a RotNet for 500 epochs with a learning rate schedule directly extracted from~\cite{chen2018semi}. We summarise the results in~\cref{tab:self-sup}.

%% Transfer Learning
\begin{table}[t]
\centering
\begin{tabular}{lccccc}
 & \multicolumn{4}{c}{Self-supervised training}\\
  & \multicolumn{4}{c}{stopped at epoch} & FT\\
	& 60 & 120 & 150 & 200 & 
\\
\hline
    Rotation accuracy  &74.1 & 83.5 & 85.9 & 86.1 & \bfseries 86.8\\
\hline
	Labelled set  & 37.6 & 40.4 & \bfseries 40.9  & 40.7 & 40.4
\\
\hline
	SSL (our method) & 46.5  & 50.2  & 50.4  &  50.7 & \bfseries 52.3

\end{tabular}
\caption{\textbf{Rotation accuracy vs Classification accuracy.} Correlation between rotation accuracy and final SSL classification accuracy on CIFAR-100 with 10 labels per class.  `FT' is fine-tuned model for 500 epochs. `Labelled' set indicates fine-tuning on label set only.}
\label{tab:self-sup}
\end{table}

Overall we found that fine-tuning on the proxy task was beneficial to the SSL algorithm. Interestingly, we note that the accuracy on the labelled set does not entirely correlate with the final SSL accuracy. While the model with best accuracy on the labelled set obtained $50.4$\% SSL accuracy, the fine-tuned model obtained an overall SSL accuracy of $52.3$\%. This also corroborates the results found in~\cref{ssec:arch-search}, better accuracy on the complete or partial set does not necessarily mean a better final SSL accuracy. We believe this result also motivates a new method of self-supervised learning evaluation using SSL.

%\begin{tabular}{|l|c|c|c|c|}
%  \hline
%   & Initial test acc & Final test acc & Upper bound test acc & Unlabeled training acc \\
%  \hline
%  CIFAR-10 & 67.5 & 84.5 & 94    & ~84 \\
%  CIFAR-100 & 38  & 47.5   & 77.34 & 47.5 \\
%  SVHN      & 81  & 96   & 97.14 & 93.2 \\
%  STL-10      & ?  & ?   & ? & ? \\
%  Pascal VOC      & ?  & ?   & ? & ? \\
%  \hline
%\end{tabular}\\
%
%\subsection{Comparison}
%
%
%\begin{tabular}{|l|c|c|c|}
%  \hline
%   Methods & SVHN & CIFAR10 & CIFAR100 \\
%  \hline
%  DGM & 36.02 $\pm$ 0.10  & - & -   \\
%  $\Gamma$-model  & - & 20.40 $\pm$ 0.47   & - \\
%  CatGAN      & -  & 19.58 $\pm$ 0.58   & - \\
%  VAT        & 24.63 & - & - \\
%  ADGM     & 22.86  & -   & - \\
%  SDGM      &16.61 $\pm$ 0.24  & -   & - \\
%  ImpGAN    & 8.11 $\pm$ 1.3  & 18.63 $\pm$ 2.32 & - \\
%  ALI       & 7.42 $\pm$ 0.65 & 17.99 $\pm$1.62  & - \\
%  $\Pi$-model  & 4.82 $\pm$ 0.17 & 12.36 $\pm$ 0.31 & 39.19 $\pm$ 0.36 \\
%  Temporal Ensembling  & 4.42 $\pm$ 0.16 & 12.16 $\pm$0.24 & 37.34 $\pm$ 0.44 \\
%  Mean Teacher	& \textbf{3.95 $\pm$ 0.19} & 12.31 $\pm$ 0.28 & - \\
%  MA-DNN~\cite{chen2018semi} & 4.21 $\pm$ 0.12 & \textbf{11.91 $\pm$ 0.22}  & \textbf{34.51$\pm$ 0.61} \\
%  \hline
%\end{tabular}\\

%% file: conclusion.tex
\section{Conclusion and discussion}
% In this paper, we have shown that current state-of-the-art SSL methods could greatly benefits from self-supervised learning. In particular we show that, to equivalent settings, self-supervision combined with SSL always provided better results.  We demonstrate that this technique also shows robustness to extreme cases with sparsely annotated data using only the dataset available.

% We also contributed a new alternative optimisation SSL algorithm. Our algorithm is particularly efficient when used as a refinement method: on a network pre-trained either with an existing SSL algorithm or when transferinf from one dataset to another. We show that it is better suited in this the latter than other SSL methods.

% Through evaluation on different architectures we showed that transfer learning can open the way to the use of deeper architectures in SSL. 

% We think that SSL could be a new and sensitive downstream task for ranking self-supervised learning methods. Although we have demonstrated our SSL here on classification, the methods are equally applicable to other tasks. As future work, we plan to apply the same technique to other tasks and datasets such as tracking or medical imaging.

% We believe that this could also help to rank self-supervision algo.
% transfer learning in SSL as a best practice for practitioner. 

%Maybe apply data augmentation on the feature level ?
%We believe it also opens the way to bigger network
%  We propose that self-supervision as well as the regime of few labelled instances per class could be used in a more systematic way to evaluate SSL algorithms.

In this paper, we have pushed the limits of SSL to work with very few annotations. We achieved this goal with several innovations. We first proposed to leverage the power of transfer learning among different tasks and self-supervision to provide a good initial feature representation for SSL. We also proposed a new SSL algorithm that can exploit well such a pre-trained representation, achieving state-of-the-art results on a large number of public benchmarks. We showed that self-supervised learning can not only benefit our proposed SSL algorithm, but also existing state-of-the-art methods. Equipped with the representation learned from self-supervised learning, we found that the performance of state-of-the-art methods can be notably boosted, especially the extreme cases with sparsely annotated data. More generally, we found that SSL could be a new and sensitive downstream task for ranking self-supervised learning methods. 
Besides working as a standalone method showing superior performance, our method can also serve as an effective refinement 
method on a network pre-trained either with an existing SSL algorithm or another dataset.

\noindent \textbf{Acknowledgments}
This work is supported by the EPSRC Programme Grant Seebibyte  EP/M013774/1, Mathworks/DTA DFR02620, and ERC IDIU-638009.

%Although we have demonstrated our SSL method here on classification, it is equally applicable to other tasks. As future work, we plan to apply the same technique to other tasks and datasets such as tracking or medical imaging.
%Compared with other SSL methods, our method is better suited with ImageNet pretrained representation.
%Through evaluation on different architectures, we think that transfer learning can open the way to the use of deeper architectures in SSL, which has not been shown much success. %As future work, we plan to apply the same technique to other tasks and datasets such as tracking or medical imaging.

% \vspace{-1.4em}
% \paragraph{\bf Acknowledgements}
% This work is supported by the EPSRC Programme Grant Seebibyte EP/M013774/1, Mathworks/DTA DFR02620, ERC 677195-IDIU and the ERC Grant ERC DFR01600.